\documentclass[letterpaper, 10 pt, conference]{ieeeconf}
\IEEEoverridecommandlockouts    

\usepackage{url}

\usepackage{color}
\usepackage{amsmath}
\usepackage{todonotes}
\usepackage{xspace}
\let\labelindent\relax
\usepackage{enumitem}
\usepackage{algorithm}
\usepackage{algpseudocode}
\usepackage{amsfonts}
\usepackage{cite}
\usepackage{subcaption}
\usepackage[hidelinks]{hyperref}
\usepackage{nameref}

\usepackage{epsfig}
\usepackage{graphicx}
\usepackage{amsmath}
\usepackage{amssymb}
\usepackage{booktabs}
\usepackage{subcaption}
\usepackage{multirow}
\usepackage[normalem]{ulem}
\hypersetup{
    colorlinks=true,
    urlcolor=magenta
    }

\usepackage{xcolor}
\definecolor{RedNew}{RGB}{255, 0, 0}
\definecolor{Blue}{RGB}{0, 0, 255}
\definecolor{DarkYellow}{RGB}{255, 213, 0}
\definecolor{DarkGreen}{RGB}{100, 213, 100}

\usepackage{hyperref}

\begin{document}

%%%%%%%%% TITLE
\title{\LARGE \bf ProxMaP: Proximal Occupancy Map Prediction\\ for Efficient Indoor Robot Navigation}

\author{Vishnu D. Sharma \and Jingxi Chen \and Pratap Tokekar
\thanks{The authors are with the University of Maryland, College Park, USA {\tt\small \{vishnuds, ianchen, tokekar\}@umd.edu}. This work is supported by the National Science Foundation under grant number 1943368 and ONR under grant number N00014-18-1-2829.}
\thanks{We thank Dr. Minghan Wei for insightful discussions and  helping with the implementation details of the baseline method.}
}

\maketitle

\begin{abstract}
   Planning a path for a mobile robot typically requires building a map (e.g., an occupancy grid) of the environment as the robot moves around. While navigating in an unknown environment, the map built by the robot online may have many as-yet-unknown regions. A conservative planner may avoid such regions taking a longer time to navigate to the goal. Instead, if a robot is able to correctly predict the occupancy in the occluded regions, the robot may navigate efficiently. We present a self-supervised occupancy prediction technique, ProxMaP, to predict the occupancy within the proximity of the robot to enable faster navigation. We show that ProxMaP generalizes well across realistic and real domains, and improves the robot navigation efficiency in simulation by \textbf{$12.40\%$} against a traditional navigation method. We share our findings and code at \url{https://raaslab.org/projects/ProxMaP}.
\end{abstract}

%%%%%%%%% BODY TEXT
\section{Introduction}
To navigate in a complex environment, a robot needs to know the map of the environment. This information can either be obtained by mapping the environment beforehand, or the robot can build a map online using the onboard sensors. Occupancy maps are often used, which provide probabilistic estimates about the free (navigable) and occupied (non-navigable) areas. These estimates can be updated as the robot gains new information while navigating.
% Online mapping reduces the deployment time and is especially suitable for navigating previously unexplored environments. 
% RGBD cameras are widely used for mobile robot sensing and the recent advancements in computer vision, fuelled by deep learning, have improved the robots' capability to understand the scene. 
Given an occupancy map, the robot can adjust its speed to navigate faster through high-confidence, free areas and slower through low-confidence, free areas so that it can stop before collision. The effective speed of the robot thus depends on the occupancy estimates.
Occlusions due to obstacles and limited field-of-view (FoV) of the robot leads to low-confidence occupancy estimates, which limit the navigation speed of the robot.

\begin{figure}
\vspace{1.5mm}
  \centering
  \begin{subfigure}{0.45\linewidth}
    % \fbox{\rule{0pt}{2in} \rule{.9\linewidth}{0pt}}
    \includegraphics[width=1.0\linewidth]{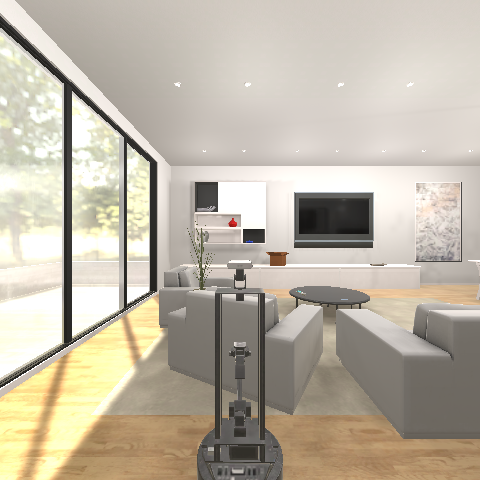}
    \caption{Third-person view of the robot in a living room}
    \label{fig:robot_eg}
  \end{subfigure}
%   \hfill
  \begin{subfigure}{0.41\linewidth}
    \includegraphics[width=1.0\linewidth]{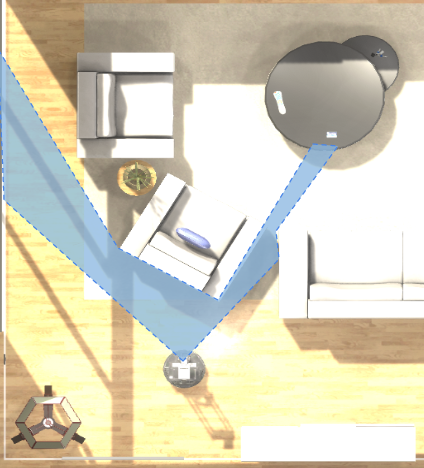}
    \caption{Top view of the robot showing visibility polygon}
    \label{fig:camera_eg}
  \end{subfigure}
  \vspace{-1mm}
  \caption{An example situation where the robot's view is limited by the obstacles (sofa blocking the view) and the camera field of view (sofa on the right is not fully visible).}
  \label{fig:real_bot}
  \vspace{-7mm}
\end{figure}

In this paper, we train a neural network to predict occupancy in the regions that are currently occluded by obstacles, as shown in Fig.~\ref{fig:real_bot}. Prior works learn to predict the occupancy map all around the robot i.e., simulating a $360^\circ$ FoV given the visible occupancy map within the current, limited FoV~\cite{ramakrishnan2020occupancy, georgakis2022uncertainty,katsumata2022map}. 
% We show this situation in Fig.~\ref{fig:fullview_vs_our}. 
Since the network is trained to predict the occupancy map all around the robot, it overfits by learning the room layouts. This happens as the network must learn to predict the occupancy information about the areas such as the back of the robot, for which the robot may not have any overlapping information in its egocentric observations. This makes the prediction task difficult to learn. Furthermore, the ground truth requires mapping the whole scene beforehand, which could make sim-to-real transfer tedious. It also means that the whole environment needs to be mapped to get the ground truth data.
% An additional challenge with this approach is that it requires mapping the whole environment in order to obtain the ground truth.

Our key insight is to simplify this problem by making predictions only about the proximity of the areas where the robot could move immediately. This setting has three-fold advantages: first, the prediction task is easier and relevant as the network needs to reason only about the immediately accessible regions (that are partly visible); second, the robot learns to predict obstacle shapes instead of learning room layouts, making it more generalizable; and third, ground truth is easier to obtain which can be obtained by moving the robot, making the approach self-supervised.

\begin{figure*}
  \centering
  \vspace{1.5mm}
  \mbox{}\hfill 
  \begin{subfigure}{0.43\linewidth}
    % \fbox{\rule{0pt}{2in} \rule{.9\linewidth}{0pt}}
    \includegraphics[width=1.0\linewidth]{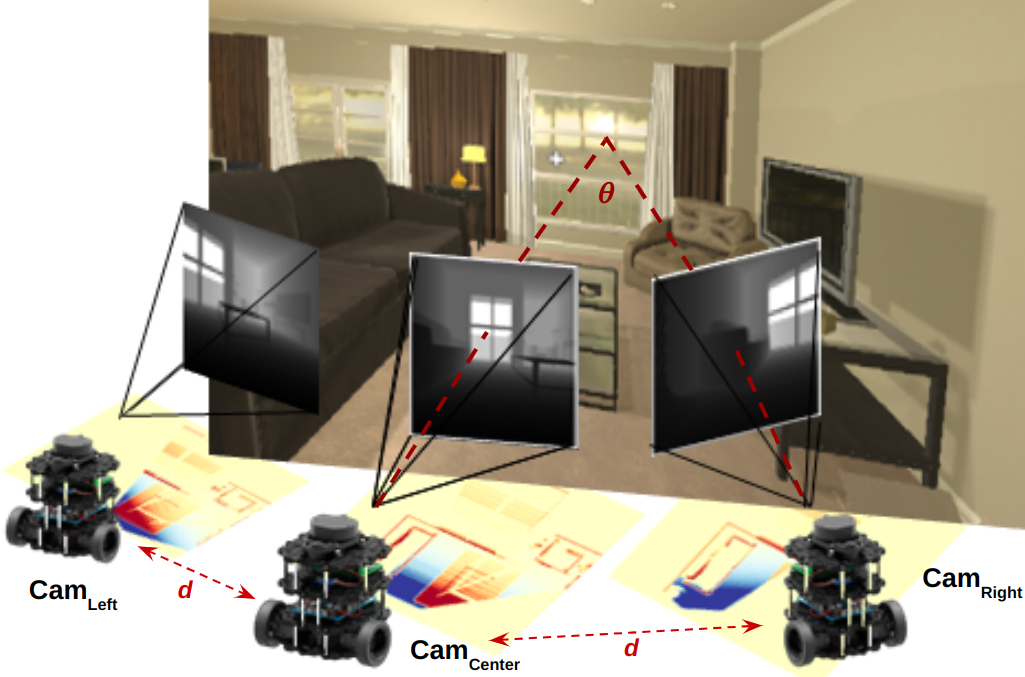}
    \caption{Movement configuration for data collection.}
    \label{fig:over_a}
  \end{subfigure}
  \hfill
  \vspace{1.5mm}
  \begin{subfigure}
  {0.45\linewidth}
    % \fbox{\rule{0pt}{2in} \rule{.9\linewidth}{0pt}}
    \includegraphics[width=1.0\linewidth]{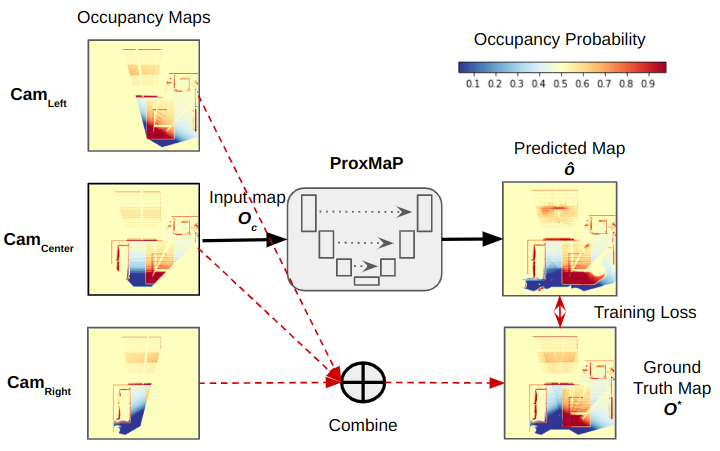}
    \caption{Training and prediction overview.}
    \label{fig:over_b}
  \end{subfigure}
  \hfill\mbox{}
  \vspace{-1mm}
  \caption{Overview of the proposed approach. The \textcolor{red}{\textbf{training}} and \textbf{inference} flows are indicated with red and black arrows, respectively. We take the input view by moving the robot to the left and right sides (\textit{$Cam_{Left}$} and \textit{$Cam_{Right}$}), looking towards the region of interest. ProxMaP makes predictions using the \textit{$Cam_{Center}$} only, and the map obtained by combining the information from the three positions acts as the ground truth. }
  \label{fig:overview}
  \vspace{-5mm}
\end{figure*}

Following are our main contributions in this work:
\begin{enumerate}
\item We present ProxMaP, a self-supervised proximal occupancy map prediction method for indoor navigation, trained on occupancy maps generated from AI2THOR~\cite{kolve2017ai2} simulator, and show that it makes accurate predictions and also generalizes well on HM3D dataset~\cite{ramakrishnan2021hm3d} without fine-tuning. 

\item We study the effect of training ProxMaP under various paradigms on prediction quality and navigation tasks, highlighting the role of training methods on occupancy map prediction tasks. 
We also present some qualitative results on real data showing that ProxMaP can be extended to real-world inputs.

\item We simulate the point goal navigation as a downstream task utilizing our method for occupancy map prediction and show that our method outperforms the baseline, non-predictive approach, relatively by $12.40\%$ in navigating faster, and even outperforms a robot with multiple cameras in the general setting.
\end{enumerate}

%-------------------------------------------------------------------------
\vspace{-1mm}
\section{Related Works}
\label{sec:related}
Mapping the environment is a standard step for autonomous navigation. The classical methods typically treat unobserved (i.e., occluded locations) as unknown. Our focus in this work is on learning to predict the occupancy values in these occluded areas. As shown by recent works, occupancy map prediction can help the robot navigate faster~\cite{katyal2021high} and in an efficient manner~\cite{elhafsi2020map}.

% Rewriting
Earlier works explored machine learning techniques for online occupancy map prediction~\cite{o2012gaussian,doherty2017bayesian}, but they require updating the model online with new observations. Recent works shifted to offline training using neural networks, treating map-to-map prediction as an inpainting task. Katyal et al.\cite{katyaloccupancy} compared ResNet, UNet, and GAN for 2D occupancy map inpainting with LiDAR data, finding that UNet outperforms the others. Subsequent works used UNet for occupancy map prediction with RGBD sensors, demonstrating improved robot navigation\cite{ramakrishnan2020occupancy,georgakis2022uncertainty,wei2021occupancy}. 
% Removed this
% Extensions to incorporate semantic information showed further benefits~\cite{georgakis2022cross}. 
Offline training for these methods requires collecting ground truth data by mapping the entire training environment, which can be time-consuming and hinder real-world deployment. Moreover, these models are trained to predict occupancy for the entire surroundings of the robot, including the scene behind for which they may lack context within the current observation, which could result in the networks memorizing room layouts, affecting their generalizability. Additionally, methods relying on historical observations for predictions~\cite{rummelhard2015conditional} face data efficiency challenges during training.

As robots can actively collect data, self-supervised methods have been successful in addressing data requirements for various robotic learning tasks~\cite{stolzle2022reconstructing, hu2021safe, khurana2022differentiable, WANG2021340, Wellhausen2019where, dhami2023prednbv}. For occupancy map prediction in indoor robot navigation, Wei et al.~\cite{wei2021occupancy} proposed a self-supervised approach using two downward-looking RGBD cameras at different heights. The network predicts the combined occupancy map from the lower camera's input without manual annotation, making it data-efficient and suitable for real robots. However, it struggles to predict edge-like obstacles and requires additional data collection for fine-tuning. Moreover, tilted cameras limit the captured information ahead compared to straight, forward-looking cameras.

To this end, we propose a self-supervised method consisting of a single, forward-looking camera to maximize the information acquisition for the navigation plane, while reducing the control effort required to collect data.  Adding two cameras to the side of the robot could further reduce this effort. We design our predictor as a classification network, which can generate sharper images compared to the regression networks, as shown later in this work. We focus on making predictions in the proximity of the robot, reducing the likelihood of memorization and improving its generalizability by using the current view as context. 

%-------------------------------------------------------------------------

\section{Approach}
\label{sec:approach}
In this work, we consider a ground robot equipped with an RGBD camera in indoor environments. Two additional views are obtained by moving the robot around as shown in Fig.~\ref{fig:over_a}. The same can also be achieved by adding extra cameras to the robot. In the following subsections, we detail the network architecture for ProxMaP, training details, and the data collection process.

\vspace{-1.0mm}
\subsection{Network Architecture and Training Details}
\label{sec:network}
\vspace{-0.5mm}
We use the occupancy map generated by $Cam_{Center}$ as input and augment it using a prediction network. Our goal is to accurately predict the occupancy information about the \textit{unknown} cells in the input map. The network uses the map generated by combining information from the three robot positions as the ground truth for training and thus learns to predict occupancy in the robot's proximity. We use  UNet~\cite{ronneberger2015u} for map prediction in ProxMaP due to its ability to perform pixel-to-pixel prediction well by sharing intermediate encodings between the encoder and decoder. We use a UNet with a 5-block encoder and a 5-block decoder. For training, we convert these maps to 3-channel images representing \textit{free}, \textit{unknown}, and \textit{occupied} regions. This is done by assigning each cell to one of the 3 classes based on its probability \textit{p}: if $p \le 0.495$, the cells are treated as \textit{free}; if $p \ge 0.505$, it is treated as \textit{occupied}; and as \textit{unknown} in rest of the cases, similar to Wei et al.~\cite{wei2021occupancy}. We train the network with cross-entropy loss, a popular choice for training classification networks.

 Since previous works have used variations of UNet for occupancy map prediction training as a regression task~\cite{wei2021occupancy, stolzle2022reconstructing} and as a generative task~\cite{katyal2019uncertainty, katyal2021high}, we also train ProxMaP with these variations. We also use UNet as the building block for these approaches with $O_c$ as input and $O^*$ as the target map (Fig.~\ref{fig:over_b}). For the regression tasks, these maps are transformed from log-odds to probability maps before training. For generative tasks, we use the UNet-based pix2pix~\cite{isola2017image} network with single and three-channel input and output pairs for regression and classification, respectively.
 
 For regression, since both input and output are probability maps, we use the KL-divergence loss function for training UNet, which simplifies to binary cross-entropy (BCE) under the assumption that each occupancy map is sampled from a multivariate Bernoulli distribution parameterized by the probability of each cell. In addition, we also train a UNet with Mean Squared Error (MSE) loss for regression. 
 For training the generative models, we use $L_{1}$ loss and $L_{GAN}$ losses as suggested by Isola et al.~\cite{isola2017image}. 
 
 In the rest of the discussion, while discussing ProxMaP's variations, we will refer to the generative classification, generative regression, and discriminative regression approaches as \textit{Class-GAN}, \textit{Reg-GAN}, and \textit{Reg-UNet}, respectively.

%-------------------------------------------------------------------------
\vspace{-1.0mm}
\subsection{Data Collection}
\label{sec:datacollection}
\vspace{-0.5mm}
We use the AI2THOR~\cite{kolve2017ai2} simulator, which provides photo-realistic scenes with depth and segmentation maps. Our setup, as shown in Fig.~\ref{fig:over_a}, includes three RGBD cameras: \textbf{$Cam_{Center}$}, positioned at the robot's height of 0.5m and location, and two additional observations from \textbf{$Cam_{Left}$} and \textbf{$Cam_{Right}$}, located at a horizontal distance of 0.3m from the original position towards left and right, respectively. Each camera is rotated by $30\deg$ to capture extra information and increase the robot's FoV. This is done to capture extra information about the scene, while also making sure that the cameras on the sides have some overlap with \textbf{$Cam_{Center}$} The rotation of the cameras virtually increases the FoV for the robot and the translation makes sure that the robot is able to learn to look around the corners rather than simply rotating at its location.

Each camera captures depth and instance segmentation images. The depth image aids in creating a 3D re-projection of the scene into point clouds, while the segmentation image identifies the ceiling (excluded from occupancy map generation) and the floor (representing the free/navigable area). The rest of the scene is considered occupied/non-navigable. The segmentation-based processing can be replaced with height-based filtering of the ceiling and floor after re-projection. All the point clouds are reprojected to a top-down view in the robot frame using appropriate rotation and translation. Maps are then limited to a $5m\times5m$ area in front of the robot and converted to $256\times256$ images to use in the network. Points belonging to obstacles increment the corresponding cell value by 1, while floor points decrement by 1. Each bin's point count is multiplied by a factor $m=0.1$ to obtain an occupancy map with log-odds. To limit log-odds values, the point count is clipped to the range $[-10,10]$. The resulting map from $Cam_{Center}$, denoted $O_c$, is the network input. The ground truth map $O^*$ is constructed as a combination of the maps from the three cameras, similar to Wei et al.~\cite{wei2021occupancy}, as follows:
\begin{equation}
  O^* = \mathrm{max}\{\mathrm {abs}(O_{c}), \mathrm {abs}(O_{l}), \mathrm {abs}(O_{r})\} \cdot \mathrm {sign}(O_{c} + O_{l} + O_{r} ),
  \label{eq:gtmap}
\end{equation}
where $O_c$, $O_l$ and $O_r$ refer to the occupancy maps generated by $Cam_{Center}$, $Cam_{Left}$, and $Cam_{Right}$, respectively. These log-odds maps are converted to probability maps before being used for network training.

AI2THOR provides different types of rooms. We use living rooms only as they have a larger size and contain more obstacles compared to others. Out of the 30 such rooms, we use the first 20 for training and validation and the rest for testing. For data collection, we divide the floor into square grids of size 0.5m and rotate the cameras by $360^\circ$ in steps of $45^\circ$. Some maps do not contain much information to predict due to the robot being close to the walls. Thus, we filter out map pairs where the number of occupied cells in $O^*$ is more than $20\%$. This process provides us with $\sim$6000 map pairs for training and $\sim$2000 pairs for testing. 
% the network. 

\section{Experiments \& Results}
\label{sec:expresults}
We report two types of results in this section. First, we present the prediction performance of the ProxMaP and its variations on our test dataset from AI2THOR. Additionally, we show prediction results on HM3D~\cite{ramakrishnan2021hm3d} to test generalizability. Then we use these networks for indoor point-goal navigation and compare them with non-predictive methods and state-of-the-art self-supervised approach~\cite{wei2021occupancy}. Finally, we present qualitative results on some real observations to highlight the potential of real-world applications. The networks were trained on a GeForce RTX 2080 GPU, with a batch size of 4 for GANs and 16 for discriminative models. Early stopping was used to avoid overfitting with the maximum number of epochs set to 300. 
% $\lambda$ was set to 100 for training GANs.

\vspace{-1.0mm}
\subsection{Occupancy Map Prediction}
\vspace{-0.5mm}
\textbf{Setup.} As our ground truth maps are generated from a limited set of observations, they may not contain the occupancy information of all the surrounding cells. Hence, we evaluate the predictions only in cells whose ground truth occupancy is known to be either occupied or free. We refer to such cells as \textit{inpainted cells}. For classification, we choose the most likely label as the output for each cell.  For regression, a cell is considered to be \textit{free} if the probability \textit{p} in this cell is lesser than $0.495$. Similarly, a cell with $p \ge 0.505$ is considered to be \textit{occupied}. The remaining cells are treated as \textit{unknown} and are not considered in the evaluations.

Prediction accuracy is a typical metric to evaluate the prediction quality. However, it may not present a clear picture of our situation due to the data imbalance caused by fewer occupied cells. Ground robots with cameras at low heights, similar to our case, are more prone to data imbalance as the robot may only observe the edges of the obstacles. Thus we also present the precision, recall, and F1-score for each class.

\begin{figure*}
\vspace{-3mm}
  \centering
\includegraphics[width=0.90\linewidth]{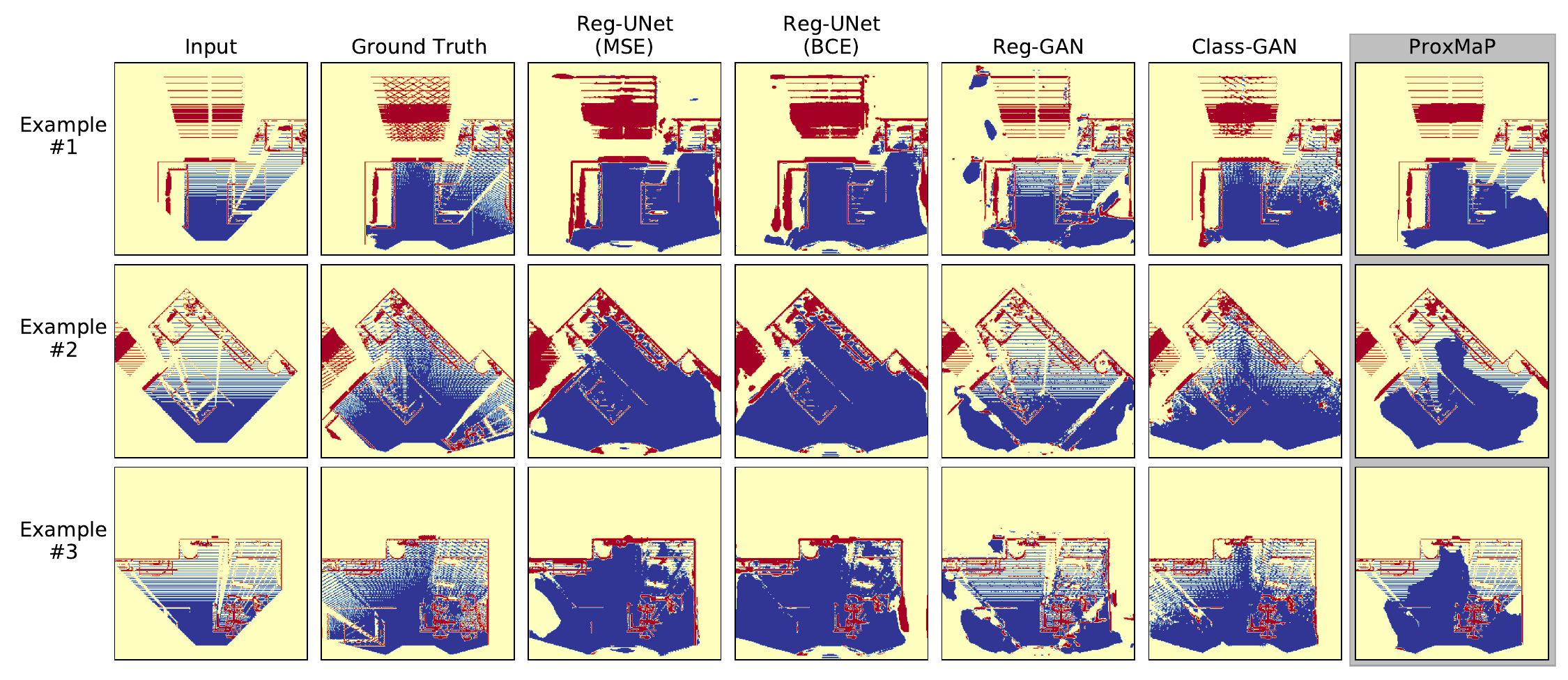}
 \vspace{-2.5mm}
  \caption{Results obtained by the proposed model over some examples (rows). Red, yellow, and green areas represent a \textcolor{RedNew}{\textbf{high}}, \textcolor{DarkYellow}{\textbf{moderate}}, and \textcolor{Blue}{\textbf{low}} chance of occupancy in an area. ProxMaP makes more accurate and precise predictions than others.}
  \label{fig:example_plots}
  \vspace{-6.5mm}
\end{figure*}

\textbf{Results.} Fig.~\ref{fig:example_plots} shows the qualitative results from ProxMaP and its variants, and Table~\ref{tab:metrics} summarizes the quantitative outcomes. The classification version of ProxMaP exhibits superior precision in predicting \textit{occupied} cells. In contrast, regression networks tend to predict surrounding areas of observed \textit{occupied} cells as \textit{occupied}, leading to higher recall but lower precision. The wider precision gap here results in a higher F1-score. The generative networks struggle due to closely mimicking patterns, including those of unknown cells. For instance, \textit{Reg-GAN} and \textit{Class-GAN} try to replicate locations of unknown cells in the ground truth, as observed in Example 3's bottom right corner. In the regression task, MSE loss outperforms BCE, which diffuses observed cells and hampers performance compared to MSE loss.

We also evaluate the generalizability of ProxMaP by testing it on similarly obtained $\sim$5000 map pairs from the HM3D dataset~\cite{ramakrishnan2021hm3d}, which contains sensor data from a realistic setting. The results are summarized in Table~\ref{tab:metrics_habitat}. We find that discriminative regression models exhibit higher accuracy than other models. However, ProxMaP continues to lead in terms of F1-score and precision for the known classes, consistent with the results in Table~\ref{tab:metrics}. All the metrics are lower than the results on AI2THOR due to differences in data sources, as expected.

In summary, classification networks perform better than regression networks on occupancy prediction. Generative approaches work better for classification than regression but one should be careful in using them as they tend to learn all patterns in the ground truth, including the undesirable ones. We present more qualitative results on our \href{https://raaslab.org/projects/ProxMaP/}{project webpage}.

\begin{table*}
\vspace{1.5mm}
\caption{Comparison across different variations of ProxMaP over living room data from AI2THOR~\cite{kolve2017ai2} simulator. Abbreviations \textit{Reg} and \textit{Class} refer to \textit{Regression} and \textit{Classification} tasks, respectively}
\vspace{-3.5mm}
\begin{center}
\begin{tabular}{ l  c  c c c  c c c  c c c } 
% \hline
\toprule
\multirow{3}*{Method} & \multicolumn{3}{c}{F1-Score} & \multicolumn{3}{c}{Precision}  & \multicolumn{3}{c}{Recall} & \multirow{3}*{Accuracy}\\ 
\cmidrule(lr){2-4} \cmidrule(lr){5-7} \cmidrule(lr){8-10} 
{} & Free & Unknown & Occupied & Free & Unknown & Occupied & Free & Unknown & Occupied & {}\\
% Method & Accuracy & Free & Unknown & Occupied & Free & Unknown & Occupied & Free & Unknown & Occupied \\
% \hline
\midrule
% \hline
% \midrule
Reg-UNet (MSE) & $82.09\%$ & $90.39\%$ & $67.08\%$ & $73.91\%$ & $96.31\%$ & $60.28\%$ & $92.32\%$ & $85.15\%$ & $\textbf{75.62\%}$ & $90.94\%$ \\
Reg-UNet (BCE) & $81.37\%$ & $89.53\%$ & $65.08\%$ & $72.03\%$ & $\textbf{96.76\%}$ & $57.49\%$ & $\textbf{93.51\%}$ & $83.30\%$ & $74.99\%$ & $90.85\%$\\
Reg-GAN & $77.90\%$ & $91.14\%$ & $67.51\%$ & $81.78\%$ & $89.57\%$ & $69.19\%$ & $74.38\%$ & $92.77\%$ & $65.91\%$ & $86.92\%$ \\
Class-GAN & $82.86\%$ & $93.08\%$ & $72.22\%$ & $82.58\%$ & $92.32\%$ & $81.14\%$ & $83.14\%$ & $93.85\%$ & $65.07\%$ & $89.71\%$  \\
\textbf{ProxMaP} & $\textbf{85.43\%}$ & $\textbf{94.19\%}$ & $\textbf{76.12\%}$ & $\textbf{87.42\%}$ & $92.38\%$ & $\textbf{88.07\%}$ & $83.52\%$ & $\textbf{96.07\%}$ & $67.02\%$ & $\textbf{92.44\%}$ \\
% \hline
\bottomrule
\end{tabular}
\end{center}
\label{tab:metrics}
\vspace{-3.5mm}
\end{table*}

\begin{table*}
\caption{Generalizability of ProxMaP and variations over Habitat-Matterport3D (HM3D)~\cite{ramakrishnan2021hm3d} dataset. Abbreviations \textit{Reg} and \textit{Class} refer to \textit{Regression} and \textit{Classification} tasks, respectively}
\vspace{-3.5mm}
\begin{center}
\begin{tabular}{ l c c c c c c c c c c } 
% \hline
\toprule
\multirow{3}*{Method} & \multicolumn{3}{c}{F1-Score} & \multicolumn{3}{c}{Precision}  & \multicolumn{3}{c}{Recall} & \multirow{3}*{Accuracy}\\ 
\cmidrule(lr){2-4} \cmidrule(lr){5-7} \cmidrule(lr){8-10} 
{} & Free & Unknown & Occupied & Free & Unknown & Occupied & Free & Unknown & Occupied & {}\\
% Method & Accuracy & Free & Unknown & Occupied & Free & Unknown & Occupied & Free & Unknown & Occupied \\
% \hline
\midrule
% \midrule
Reg-UNet (MSE) & $79.99\%$ & $85.91\%$ & $56.33\%$ & $72.24\%$ & $93.52\%$ & $44.48\%$ & $89.59\%$ & $79.45\%$ & $76.79\%$ & $90.08\%$ \\
Reg-UNet (BCE) & $78.70\%$ & $84.40\%$ & $49.92\%$ & $69.70\%$ & $\textbf{94.84\%}$ & $36.96\%$ & $\textbf{90.37\%}$ & $76.02\%$ & $\textbf{76.91\%}$ & $\textbf{90.81\%}$ \\
Reg-GAN & $75.05\%$ & $88.25\%$ & $56.21\%$ & $82.38\%$ & $85.42\%$ & $51.17\%$ & $68.92\%$ & $91.28\%$ & $62.36\%$ & $83.55\%$ \\
Class-GAN & $80.27\%$ & $90.04\%$ & $70.81\%$  & $81.65\%$ & $89.04\%$ & $76.01\%$ & $78.93\%$ & $91.05\%$ & $66.28\%$ & $86.41\%$ \\
\textbf{ProxMaP} & $\textbf{81.50\%}$ & $\textbf{90.91\%}$ & $\textbf{74.11\%}$ & $\textbf{84.32\%}$ & $89.07\%$ & $\textbf{84.22\%}$ & $78.86\%$ & $\textbf{92.83\%}$ & $66.16\%$ & $87.54\%$  \\
% \hline
% \hline
\bottomrule
\end{tabular}
\end{center}
\label{tab:metrics_habitat}
\vspace{-3mm}
\end{table*}

\vspace{-1.0mm}
\subsection{Navigation Performance}
\vspace{-0.5mm}
\textbf{Setup.} We use the occupancy map prediction for point-goal navigation in an unknown environment by repeated prediction and planning over the path. We also compare the discriminative and generative variation here and use only \textit{Reg-UNet(MSE)} as the discriminative regression method as it performs better than \textit{Reg-UNet(BCE)}. We compare these methods against an approach relying solely on the input map $O_c$, referred to as the \textit{baseline}, and the method proposed by Wei et al.~\cite{wei2021occupancy}, but trained with MSE loss instead of BCE as the network trained with  BCE loss (proposed by the authors) did not perform well. We obtained an F1-score of $44.01\%$ with BCE loss as compared to the F1-score of $47.61\%$ for the network trained with MSE loss. Additionally, we compare these methods against the setting where the robot is equipped with cameras on the sides, i.e., when $Cam_{Left}$ and $Cam_{Right}$ directly provide the inputs (thereby not requiring any prediction) about the scene (referred to as \textit{3-Cameras}).

In each trial, we randomly generate a start location, a destination, and initial yaw for the robot. At each step, the robot generates an occupancy map using the RGBD camera and maintains a global occupancy map in its memory. We utilize perceived height to filter out obstacles above the robot's height, avoiding reliance on ground truth semantic maps. We assume the robot knows its global location and yaw without error. A prediction network augments the occupancy map and the global map is updated using the same method as for ground truth generation (Eq.~\ref{eq:gtmap}). This global map is used as the cost map for path planning with Dijkstra's algorithm to find the shortest path to the destination. The robot navigates to the next prescribed waypoint, moving in 20cm steps if it is within $1^{\circ}$ of the robot's line of sight. Otherwise, the robot rotates to face the waypoint before moving. The room is discretized into a grid of square cells with sides of 0.2m. The simulation ends when the goal is within one diagonal cell at most or after the robot has moved and rotated $S_{max}$=100 times, typically sufficient to reach the goal.

For path planning, the sensed occupied and free areas are given preference over the predicted counterparts in the maps, and the costs are accordingly weighted in the cost maps. We keep the speed of the robot proportional to its confidence about the free cells on the predicted path. Thus the robot moves faster when it is confident about not colliding with an obstacle on the path. If there are possible obstacles on the path, it should keep its speed lower to be able to stop before a collision. The cost of the traversal is equal to the time taken to reach the goal.

To measure the navigation efficiency in terms of time, we use \emph{Success weighted by (normalized inverse) Completion Time}~\cite{yokoyama2021success} as the metric, which is defined as 
$SCT = \frac{1}{N} \sum_{i=1}^N S_i \frac{l_i}{max(p_i, l_i)}$
where, $N$ is the number of test episodes, $S_i$ a binary variable indicating success in the $i^{th}$ episode, $l_i$ is the traversal time on the shortest path between the source and destination, and $p_i$ is the measured traversal time by the robot in this episode. In our case, $S_i$ is 1 only when the robot is within one cell away, the number of total steps does not exceed $S_{max}$, and the simulator does not run into an error during simulation. Here, $l_i$ is found using simulation on the map generated using all the reachable positions in the environment. This map does not depend on sensing and thus is not updated during an episode. Here, $p_i$ is calculated as the time taken for traversal by the robot. A higher value of SCT is preferred, as it indicates a higher success rate and smaller difference in $p_i$ and $l_i$.

\begin{table}
\vspace{1.5mm}
\caption{SCT performance across different living rooms}
\vspace{-3.0mm}
\begin{center}
\begin{tabular}{ l c} 
% \hline
\toprule
Method & SCT \\
% \hline
% \hline
\midrule
Baseline (no prediction)                 &   0.589\\
Wei et al.~\cite{wei2021occupancy} (MSE)  &   0.568\\
Reg-UNet (MSE)    &   0.629\\
% \textbf{Regression GAN}               &   \textbf{0.591}\\
Reg-GAN               &   0.592 \\
Class-GAN    &   0.632 \\
% \hline
\midrule
\textbf{ProxMaP}   &   \textbf{0.662} \\
3-Cameras (no prediction)           &   0.648\\
% \hline
\bottomrule
\end{tabular}

\end{center}
\label{tab:nav_metrics}
\vspace{-9.0mm}
\end{table}

\textbf{Results.} First, we compare the navigation efficiency by simulating navigation in different living rooms (\textit{FloorPlan221-227}). We report SCT over $N$=$102$ episodes in these rooms in Table~\ref{tab:nav_metrics}. Most of these rooms are small and have a few situations where the robot needs to choose between multiple paths. We observe that the classification models outperform the others, reaching a maximum relative benefit of $12.39\%$ against the baseline approach. ProxMaP even outperforms the 3-cameras method by a relative margin of $2.16\%$. This happens because the prediction can also fill the gaps in the map that the 3-camera setup can not observe, resulting in a higher navigation speed over the path. The prediction method by Wei et al.~\cite{wei2021occupancy}, which relies on a higher-lower camera setup, fails to outperform the baseline in this situation. Other ProxMaP variations also outperform the baseline, highlighting the benefit of making predictions in proximity. These variations however fall short in comparison to the 3-camera setup. 

% \vspace{-0.1mm}
To study the navigation in a more complex situation with a higher emphasis on decision-making, we use a modified version of the living room $FloorPlan227$ 
% (Fig.~\ref{fig:fp227_orig}) 
due to its larger size and higher number of obstacles, and therefore, more possible paths to the goals. We remove an armchair to present more path options to the robot.
We find that over $N$=$100$ episodes in this setting, ProxMaP achieves a relative improvement in SCT by $8.62\%$ over the \textit{Baseline} and $7.2\%$ improvement over Wei et al.~\cite{wei2021occupancy}. In many cases, an erroneous prediction may discourage the robot from moving on a shorter path to the goal, resulting in longer navigation time. 
A summary of these results is presented on our \href{https://raaslab.org/projects/ProxMaP}{project webpage}.

Overall, the classification approaches perform better in both prediction and navigation, compared to the regression counterparts. Their precise predictions and generalizability make them further suitable for occupancy map prediction.

\vspace{-1.0mm}
\subsection{Predictions on Real Data}
\begin{figure}[ht!]
\vspace{-3.0mm}
  \centering
\includegraphics[width=0.90\linewidth]{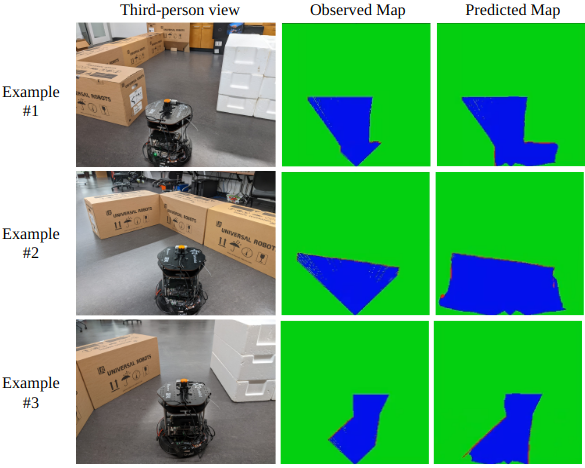}
%   \caption{Some model predictions/outputs and the corresponding input and ground truth maps.}
  \vspace{-1.0mm}
  \caption{Prediction by ProxMaP over real-world inputs.}
  \label{fig:real_results}
  \vspace{-3.0mm}
\end{figure}

\vspace{-0.5mm}
 We test ProxMaP on real data using a TurtleBot2 robot 
 equipped with a Hokuyo 2D laser scanner. The FoV of the 
 scanner is limited to $90^\circ$ and its readings are used to generate the occupancy maps. We use ProxMaP over these maps to make predictions. Fig.~\ref{fig:real_results} shows the third-person view of the robot in a maze built in our lab and the qualitative results for some interesting situations. In these maps, the \textcolor{DarkGreen}{unknown}, \textcolor{blue}{free}, and \textcolor{red}{occupied} regions are shown in green, blue, and red, respectively. Similar to our training conditions, the scanner at a low height can only see the edges of the obstacles. In these situations, we show that even when a part of the obstacle is visible, ProxMaP can estimate its shape very well, while also predicting the nearby free regions. We perform these predictions offline due to the prediction latency but will explore the real-time inference tools in future work. 

\vspace{-1.0mm}
\section{Conclusion}
\vspace{-0.5mm}
\label{sec:conclusion}
We presented ProxMaP, a self-supervised method for predicting occupancy maps in the robot's proximity to aid navigation in the presence of obstacles. Our approach demonstrates higher prediction precision, generalizability to realistic and real inputs, and improved robot navigation time by adjusting robot speed based on observed and predicted information. Furthermore, ProxMaP shows the potential to outperform equivalent multi-camera setups. We also explored different variations of ProxMaP, finding that classification-based approaches yield superior predictions, both qualitatively and quantitatively, compared to regression-based methods, and also enable the robot to navigate faster by inferring about nearby regions occluded by obstacles.

The proposed method can be further extended to study the effect of different placements of the additional camera on robot navigation. The method may also benefit by leveraging other sensor and input modalities for prediction and planning. Our preliminary studies on including semantic information show promising results and will be explored in future work. 
%-------------------------------------------------------------------------

{\small
\bibliographystyle{IEEEtran}
\bibliography{bibfile}
}

\end{document}